%% file: main.tex
\documentclass[letterpaper]{article} 
\usepackage[]{aaai2026}  
\usepackage{times}  
\usepackage{helvet}  
\usepackage{courier}  
\usepackage[hyphens]{url}  
\usepackage{graphicx} 
\usepackage{natbib}  
\usepackage{caption} 
\usepackage{algorithm}
\usepackage{algorithmic}
\usepackage{microtype}
\usepackage{subfigure}
\usepackage{amsmath}
\usepackage{amssymb}
\usepackage{mathtools}
\usepackage{amsthm}
\usepackage{mdframed}
\usepackage[capitalize,noabbrev]{cleveref}
\usepackage[textsize=tiny]{todonotes}
\usepackage{multirow}
\usepackage{enumitem}
\usepackage{booktabs}

\newtheorem{definition}{Definition}

\newmdtheoremenv[
    linecolor=black,  
    linewidth=1pt,    
    leftmargin=0pt,  
    rightmargin=0pt, 
    innertopmargin=3pt,
    innerbottommargin=3pt,
    innerleftmargin=3pt, 
    innerrightmargin=3pt 
]{blackboxtheorem}{Theorem}

\usepackage{newfloat}
\usepackage{listings}
\DeclareCaptionStyle{ruled}{labelfont=normalfont,labelsep=colon,strut=off} 
\floatstyle{ruled}
\newfloat{listing}{tb}{lst}{}
\floatname{listing}{Listing}
\title{DimINO: Dimension-Informed Neural Operator Learning}
\author{
    Yichen~Song \\
    Yalun~Wu \\
    Yunbo~Wang \\
    Xiaokang~Yang
}
\affiliations{
    \textsuperscript{\rm 1}MoE Key Lab of Artificial Intelligence, AI Institute, Shanghai Jiao Tong University

}

\usepackage{bibentry}

\begin{document}

\maketitle

\newcommand{\yb}[1]{{\color{blue}{[wyb: #1]}}}
\newcommand{\new}[1]{{\color{red!60!black}{[newly added: #1]}}}

\input{contents/0_abs}
\input{contents/1_intro}
\input{contents/4_related}

\input{contents/2_method}
\input{contents/3_exp}
\input{contents/5_concl}

\bibliography{aaai2026}

\end{document}

%% file: contents/0_abs.tex
\begin{abstract}
In computational physics, a longstanding challenge lies in finding numerical solutions to partial differential equations (PDEs). Recently, research attention has increasingly focused on Neural Operator methods, which are notable for their ability to approximate \textit{operators}---mappings between functions. Although neural operators benefit from a universal approximation theorem, achieving reliable error bounds often necessitates large model architectures, such as deep stacks of Fourier layers. This raises a natural question: \textit{Can we design lightweight models without sacrificing generalization?} To address this, we introduce \textbf{DimINO} (Dimension-Informed Neural Operators), a framework inspired by dimensional analysis. DimINO incorporates two key components, \textbf{DimNorm} and a \textbf{redimensionalization} operation, which can be seamlessly integrated into existing neural operator architectures. These components enhance the model’s ability to generalize across datasets with varying physical parameters. Theoretically, we establish a universal approximation theorem for DimINO and prove that it satisfies a critical property we term \textbf{Similar Transformation Invariance} (STI). Empirically, DimINO achieves up to $\textbf{76.3\%}$ performance gain on PDE datasets while exhibiting clear evidence of the STI property.
\end{abstract}

%% file: contents/1_intro.tex
\section{Introduction}

Neural Operator methods have gained significant attention in AI for Science due to their remarkable ability to approximate operators (\textit{i.e.}, mappings between infinite-dimensional function spaces), where traditional neural networks often fall short.
Despite their promise, recent studies on Neural Operators~\citep{FFNO, GNOT, takamoto2022pdebench} have revealed a key limitation: their performance tends to degrade when applied across diverse physical regimes (\textit{e.g.}, fluid dynamics under varying viscosities).
This highlights the need for models that are not only data-driven but also grounded in physical principles.

Our work begins with a simple yet profound insight: \textit{Neural Operators that incorporate dimensional information can better learn and generalize across PDE systems.} 
%
This intuition draws from classical dimensional analysis: in physical systems, quantities with different units cannot be directly added, and meaningful interactions typically occur via unit-consistent operations such as multiplication or normalization. However, most existing models ignore this constraint, treating all inputs as dimensionless numerical fields.

To address this gap, we propose the Dimension-Informed Neural Operator (DimINO), a framework that explicitly incorporates physical quantities and their dimensions into operator learning.
%
%
Specifically, we design two plug-and-play modules: \textbf{DimNorm} and \textbf{ReDimensionalization}, both of which can be seamlessly integrated into existing Neural Operator models without increasing parameter count.
DimNorm normalizes the input data by incorporating dimensionless numbers, enabling the Neural Operator to process inputs governed by the dimensionless form of the underlying PDE. ReDimensionalization then maps the dimensionless outputs back to physical fields with proper dimensions, ensuring consistency with the original physical quantities.

The incorporation of dimensional information in DimINO is NOT equivalent to simply appending a few \textit{dimensionless numbers} to the model inputs.
For instance, in FNO-based models, replacing DimNorm with the concatenation of dimensionless numbers fails to bring improvement.
Instead, DimINO provides a principled mechanism for coupling field variables with physical constants in a way that respects their dimensional relationships. In tasks where no meaningful physical constants are defined, DimNorm gracefully reduces to a standard LayerNorm~\citep{ba2016layer}. 

Beyond empirical design, DimINO is also grounded in theory. We show that it aligns with a core principle from dimensional analysis: \textbf{Similar Transformation Invariance} (STI)---the property that solutions to physical PDEs remain invariant under rescaling of units. We theoretically prove that DimINO models exhibit STI, thereby ensuring that they generalize in a physically consistent manner.

We evaluate DimINO on classic PDE simulation benchmarks, including \textit{Advection}, \textit{Burgers}, \textit{Diffusion-Reaction}, and \textit{Navier-Stokes}. Experimental results not only demonstrate its improved performance across diverse regimes but also empirically confirm its STI property, highlighting the generalization ability of the approach.

%% file: contents/4_related.tex
\section{Related Work}

\subsection{Numerical PDE Solvers}
Traditional approaches for solving PDEs rely on numerical techniques such as the finite element method (FEM), finite difference method (FDM), finite volume method (FVM), and pseudo-spectral methods~\citep{grossmann2007numerical,solin2005partial,ciarlet2002finite,courant1967partial,cooley1969fast,gottlieb1977numerical,fornberg1998practical,kopriva2009implementing} to solve PDEs.
These methods discretize the spatial domain, where achieving higher accuracy requires finer discretization, often leading to significantly increased computational costs. To reduce these costs, simplified models, such as Reynolds-Averaged Navier-Stokes (RANS) and Large Eddy Simulations (LES), have been developed to approximate complex dynamics more efficiently.

\subsection{Learning-Based PDE Solvers}

Recent advances in machine learning have introduced promising alternatives for accelerating scientific simulations. Among them, Physics-Informed Neural Networks (PINNs)~\citep{PINN,lu2021deepxde,karniadakis2021physics} represent an early class of PDE learners. These methods embed the governing equations directly into the loss function~\citep{yu2018deep,wang2022and,wang2021understanding}.
However, PINN-based approaches often suffer from poor generalization across varying boundaries and initial conditions. Moreover, they require explicit knowledge of the underlying physical system, which limits their applicability in data-driven scenarios or systems with partially known physics.
%

Neural operator learning aims to build more flexible and generalizable PDE solvers, with two main foundational branches: DeepONet~\citep{DeepONet} and Fourier Neural Operator (FNO)~\citep{FNO}.
DeepONet leverages separate branch and trunk networks to encode input functions and evaluation coordinates, grounded in the universal approximation theorem~\citep{chen1995universal}.
In contrast, FNO-based methods~\citep{FFNO,TFNO,xiong2022koopman,wen2022u,ashiqur2022u,raonic2024convolutional} formulate operator learning through spectral convolution by applying discrete Fourier transforms to model kernel integral operators. FNO achieves a favorable cost-accuracy trade-off due to its quasi-linear complexity with respect to resolution.
Variants such as F-FNO~\citep{FFNO} reduce parameter count by applying 1D Fourier transforms along individual axes, while T-FNO~\citep{TFNO} adopts tensor decomposition techniques (\textit{e.g.}, Tucker factorization) to improve efficiency.

%

Another promising line of work introduces attention-based operator learners~\citep{cao2021choose,LSM,GNOT} inspired by the success of Transformers~\citep{vaswani2017attention}. 
%
LSM~\citep{LSM} uses sine functions to approximate any function guaranteed by the theorem of Convergence of Trigonometric Approximation~\citep{dyachenko1995rate}, and adopts a U-Net architecture~\citep{ronneberger2015u} to handle multi-scale structures.
GNOT~\citep{GNOT} introduces a heterogeneous normalized attention layer to handle varying grid structures and inputs.
While both LSM and GNOT show promising empirical results, it remains unclear whether they possess the same theoretical universality guarantees for operator learning as DeepONet and FNO.

%% file: contents/2_method.tex
\section{Dimension-Informed Neural Operators}

\begin{figure*}[t]
    \centering
    \includegraphics[width=0.9\linewidth]{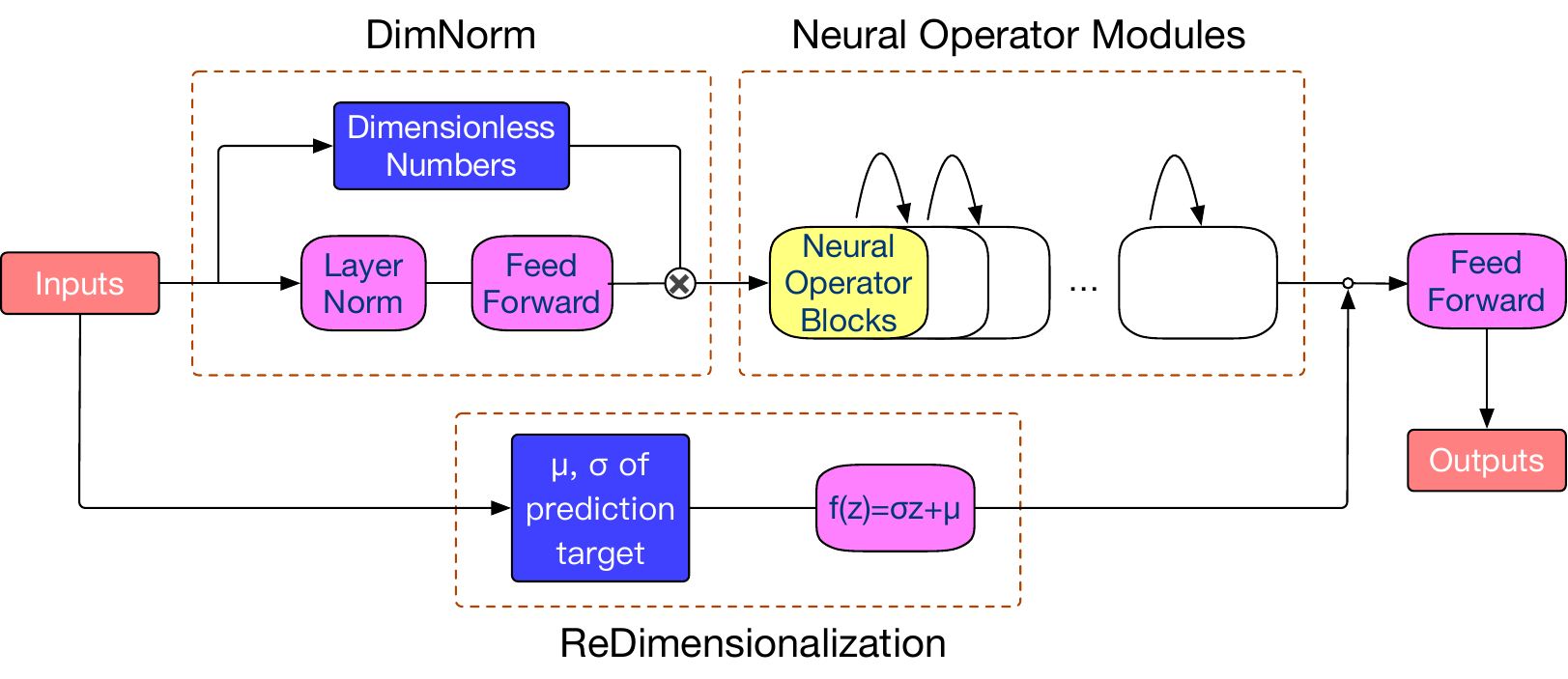}
    \vspace{-10pt}
    \caption{\textbf{An illustration of DimINO.} The input variables are first normalized using a LayerNorm operation, while the dimensionless numbers of the PDE system would be calculated and transformed by a Feed-Forward Block. These are then applied to the normalized input fields via channel-wise multiplication, enabling dimension-informed interactions. The resulting features are passed through a sequence of standard Neural Operator blocks. Finally, the output is projected to match the scale of the prediction target and refined by another Feedforward Block to produce the final outputs.}
  \label{fig:DimNorm}
  \vspace{-5pt}
\end{figure*}

In this section, we present the technical details of DimINO, a simple yet effective approach that can be seamlessly integrated into various neural operator-based PDE solvers.
We begin by introducing the core intuition and theoretical foundation behind DimINO, followed by the design of DimNorm, a novel and lightweight neural module inspired by principles of dimensional analysis.
Finally, we provide a theoretical analysis of the proposed method.

\subsection{Dimensional Analysis of Numerical PDE Solutions}
\label{sec:DimAware}

In operator learning, two key properties are widely studied: (1) \textit{universal approximation of operators}, as guaranteed by models like DeepONet, and (2) \textit{resolution invariance}, a desirable property for filter-based models such as FNO~\citep{FNO} and CNO~\citep{raonic2024convolutional}.

We argue that dimensional analysis, a classic tool in physics, should be a guiding principle in neural operator design. 
%
In the context of solving PDEs, considering dimensional information allows the model to explicitly encode relationships and interactions between different physical quantities. For instance, in many physical systems, variables are often related through products and ratios that have specific dimensional properties. 
Traditional AI models often overlook such relationships, which may result in physically invalid representations and degraded performance
By embedding dimensional information into the model, we ensure that it respects the underlying physical laws of the system, leading to improved accuracy and generalization.

To illustrate this point, consider a common prediction task in physics: forecasting the future state of a physical field given its current state. This process is governed by a dynamical equation of the form:
\begin{equation}
    \frac{d}{dt}  \mathbf{u}(\mathbf{x},t) = \mathcal{G} \mathbf{u}(\mathbf{x},t),
\end{equation}
where $\mathcal{G}$ is the time-evolution operator. In practice, $\mathcal{G}$ often consists of sums of product terms, with each term combining physical quantities of different dimensions. Despite this, all resulting terms share the same dimension as $\frac{d}{dt} \mathbf{u}(\mathbf{x},t) $, ensuring physical consistency.
Let us consider the Navier-Stokes equations~\citep{temam2001navier}, which govern fluid motion:
\begin{equation}
    \rho \frac{d}{dt} \boldsymbol{v} = - \rho \boldsymbol{v}\cdot \nabla \boldsymbol{v} -\nabla p + \mu \nabla^2 \boldsymbol{v} + \rho \boldsymbol{f},
\label{eqn:ns-eq}
\end{equation}
where $\rho$ is the fluid density,  $\boldsymbol{v}$ is the velocity field, $p$ is the pressure field, $\mu$ is the dynamic viscosity, and $\boldsymbol{f}$ is the external force field.
Each term on the right-hand side represents a different physical process (advection, pressure gradient, diffusion, and external forcing), yet all terms are dimensionally consistent with the left-hand side, representing the change rate of momentum per unit volume. 
%


However, most neural operator models like FNO or CNO treat inputs indiscriminately---linearly combining velocity fields, pressure fields, and even dimensioned constants without regard for their units. From a physics perspective, this is deeply unsatisfying and contradicts the fundamental rule that quantities of different dimensions must not be directly added or compared. Our design, DimINO, addresses this inconsistency by embedding dimensional structure into the network architecture, as will be detailed in the following section.



\subsection{DimNorm \& Redimensionalization: Aligning Scales of Latent Spaces with Dimensionless Numbers}
\label{sec:DimNorm}

The input to a typical neural operator includes three components: field variables, grids, and global constants. 
DimINO is designed for scenarios involving multiple physical quantities with distinct units, which is common in applications like computational fluid dynamics (CFD), where inputs may include velocity, pressure, and external force fields, along with global constants such as viscosity or the prediction time interval.
When complex geometries are involved, grid structures or volume fractions must also be included in the inputs.


However, most neural operator benchmarks adopt simplified settings, often fixing certain quantities and treating others as constants.
For example, FNO~\cite{FNO} evaluates only on vorticity fields (TorusLi), and PDEBench~\cite{takamoto2022pdebench} treats each combination of constants as a separate task.
In contrast, DimINO supports learning over general, multi-variable PDE scenarios, making dimensional consistency critical.
We formalize DimINO as follows:

\begin{definition}[DimINO]\label{def:DimINO}
Let $\mathrm{LayerNorm}(\cdot)$ denote standard layer normalization, and $\mathrm{DimGate}(\mathbf{c}, \mathbf{x})$ represent a soft gating mechanism modulated by dimensionless numbers $\mathbf{c}$ applied to input $\mathbf{x}$.


Specifically, given $\mathbf{x} \in \mathbb{R}^{n}$ (with $n$ channels), and $\mathbf{c} \in \mathbb{R}^{m}$, we expand $\mathbf{c}$ to $\mathbf{c}' \in \mathbb{R}^{n}$ as follows: Let $l = \lfloor (1 - \gamma)n/m \rfloor$. Then, for $i \in [0, ml-1]$,
$\mathbf{c}'[i] = \mathbf{c}[\lfloor i / l \rfloor]$, and the remaining $\gamma \cdot n$ dimensions are filled with ones (skip connections).
Let $\phi_{dim}(\cdot)$ denote the redimensionalization function.




The full DimINO pipeline is:
\begin{equation}\label{eqndef:DimINO}
    \begin{split}
        \mathrm{DimINO}(\mathbf{u}) &= \mathrm{FFW}_{post} \circ \phi_{dim} \circ \mathrm{NeuralOp} \\ 
        &\circ \ \mathrm{DimGate}(\mathbf{c}(\mathbf{u}), \ \cdot\ )  \circ \mathrm{FFW}_{pre} \\ 
        &\circ \ \mathrm{LayerNorm} (\mathbf{u}).
    \end{split}
\end{equation}
\end{definition}

Figure \ref{fig:DimNorm} shows an overview of our architecture design. The core of DimINO lies in the \textbf{DimNorm} module, which extracts dimensionless numbers, normalizes the inputs, and uses a feedforward network to project into a higher-dimensional latent space. These latent fields are then modulated via the \textbf{DimGate}, enabling explicit injection of dimensionless physical priors.
On the output side, as the neural operator processes dimensionless inputs, its output must also be dimensionless. To recover predictions in physical units, we design the \textbf{ReDimenisonalization} operation $\phi_{dim}(\cdot)$, to multiply the dimensions back to the output channels, so that the Neural Operator could return quantities with dimensions.

\subsection{Similar Transformation Invariance}
\label{sec:STI}

Similar Transformation Invariance is a fundamental principle in dimensional analysis when studying physical PDE systems. Typically, the analysis starts with \textit{nondimensionalization}, aligning all variables with characteristic scales.
This corresponds to the DimNorm operation in our framework. The final step involves restoring physical units to the target variables, which corresponds to the redimensionalization operation. 
A similar transformation preserves the dimensionless numbers of a system while scaling geometric quantities proportionally. Under such a transformation, the solution to the transformed PDE (analytical, numerical, or experimental) can be mapped back to the original system via inverse scaling.

We illustrate this concept using the Navier-Stokes equations for incompressible Newtonian fluids:
\begin{equation}
    \rho (\frac{d}{dt} + \boldsymbol{v}\cdot \nabla) \boldsymbol{v} = -\nabla p + \mu \nabla^2 \boldsymbol{v} + \rho \boldsymbol{f}.
\end{equation}
Let $L$, $U$, $t_0$, and $f_0$ denote the characteristic length, velocity, time, and force scale. We have dimensionless variables:
\begin{equation}
    \boldsymbol{v'}=\frac{\boldsymbol{v}}{U}, \boldsymbol{r'} = \frac{\boldsymbol{r}}{L}, t' = \frac{t}{t_0}, p' = \frac{p}{\rho U^2}, f' = \frac{f}{f_0}.
\end{equation}
Substituting the nondimensional variables into the original PDE yields the following dimensionless form, governed by the Reynolds number $Re = \frac{\rho U L}{\mu}$, the Strauhal number $St = \frac{t_0 U}{L}$, and the Froude number $Fr = \frac{U}{\sqrt{f_0L}}$:
\begin{equation}
    \frac{1}{St} (\frac{d}{dt'} + \boldsymbol{v}\cdot \nabla') \boldsymbol{v'} = -\nabla' p' + \frac{1}{Re} \nabla'^2 \boldsymbol{v'} + \frac{1}{Fr^2} \boldsymbol{f'}.
\end{equation}

This transformation shows that the PDE can be rewritten purely in terms of dimensionless groups. Thus, if a neural operator respects similar transformation invariance, it can generalize across physically equivalent but differently scaled systems.
We further validate this property through experiments on the TorusVisForce dataset~\cite{FFNO}. See the next section for details.

\subsection{Theoretical Analysis for DimINO's Universal Approximation Theorem and Similar Transformation Invariance}

\begin{blackboxtheorem}[DimINO Universal Approximation]

Let $G_{\Delta t}$ be the exact solution operator for a PDE system:
\begin{equation}\label{eqn:pde}
\frac{\partial \mathbf{u}}{\partial t} = \mathbf{f_0}(\partial_x, \mathbf{u}).
\end{equation}
After nondimensionalization, it transforms into:
\begin{equation}\label{eqn:nondim}
\frac{\partial \mathbf{u}^*}{\partial t} = \mathbf{f_1}(\partial_x, \mathbf{u}^*, \tilde{\mathbf{c}}) = \mathbf{f}(\partial_x, \mathbf{g}(\mathbf{u}^*, \tilde{\mathbf{c}})),
\end{equation}
where $\tilde{\mathbf{c}}$ is a vector of dimensionless numbers.

Here we assume that the ground truth Evolutionary Operator of time interval $\Delta t$ is $\mathcal{G}^\dagger_{\Delta t}(\mathbf{u}|_{t_0}) = \mathbf{u_{pred}}|_{t_0+\Delta t}: \mathcal{A}\rightarrow\mathcal{U}$, where $\mathcal{A}, \ \mathcal{U}$ are respectively the Banach spaces of functions of the input domain and output domain.

For any physical PDE system, let the characteristic quantities be $\mathbf{e_0}$, and the nondimensionalized quantities are $\mathbf{h_{e_0}}(\mathbf{u})$. 

$\mathbf{u} = \mathbf{h_{e_0}}(\mathbf{u}) \mathbf{u}^*$ is the nondimensionalized variables, and $\tilde{\mathbf{c}}(\mathbf{u}) = \mathbf{c}(\mathbf{h_{e_0}}(\mathbf{u}))$ is the vector of dimensionless numbers. 

Then for any $\epsilon > 0$, there exists a DimINO model $\mathcal{G}_\theta$ such that, given any $\mathcal{K}\subset \mathcal{A} $ compact:
\begin{equation}\label{neq:operator}
d(\mathcal{G}_\theta, \mathcal{G}_{\Delta t}) = \text{sup}_{\mathbf{u}_0\in  \mathcal{K}} \|\mathcal{G}_\theta(\mathbf{u}_0) - \mathcal{G}_{\Delta t}(\mathbf{u}_0)\|_{L^2} < \epsilon,
\end{equation}
where $\theta$ is the parameter of the Neural Operator. 
\end{blackboxtheorem}

\begin{proof}
DimINO decomposes the operator approximation into three components: preprocessing $\Phi_{\text{pre}, \theta_0}$, neural operator block $\Phi_{\text{NeuralOp},\theta_1}$, and postprocessing $\Phi_{\text{post}, \theta_2}$:
\begin{equation}
\mathcal{G}_\theta = \Phi_{\text{post}, \theta_2} \circ \Phi_{\text{NeuralOp},\theta_1} \circ \Phi_{\text{pre}, \theta_0}.
\end{equation}

Since Neural Operator blocks can degenerate into universal function approximators (such as MLPs), functional composition can be considered a special case of operator learning.
Therefore, there exists a subset of $\mathbf{NeuralOp}$ blocks that can be decomposed into two components: the first approximates a general function, while the second approximates an operator.

We assume that, through Layer Normalization and dimensional analysis, we obtain the normalized field $\mathbf{u}^*$ and the nondimensionalized coefficients $\tilde{\mathbf{c}}(\mathbf{u})$. Under this assumption, our proof sketch proceeds as follows:

We propose that each DimINO Neural Operator Block can be viewed as comprising two distinct stages. The first stage approximates the function $\mathbf{g}(\mathbf{u}^*, \tilde{\mathbf{c}}))$, while the second stage approximates the operator $\mathcal{G}^{\dagger*}_{\Delta t, \tilde{\mathbf{c}}}(\mathbf{u}^*|_{t_0}) = \mathbf{u_{pred}}^*|_{t_0+\Delta t}$, which serves as regression target of the second part of the Neural Operator module.

Formally, we assume that the $\mathrm{NeuralOp}$ block defined in Eq. \eqref{def:DimINO} can be decomposed as: $\mathrm{NeuralOp}(\mathbf{u}) = \mathrm{NeuralOp_1} \circ F_{\alpha} \circ \mathbf{u}$.
Given $\mathbf{u}^* = \mathrm{LayerNorm}(\mathbf{u})$, we aim to show that the function $\mathbf{g}$ can be well approximated by $F_{\alpha} \circ \mathrm{DimGate}(\mathbf{c}(\mathbf{u}), \ \cdot\ )  \circ \mathrm{FFW}_\text{pre} $. 
This follows from the universal approximation properties of neural networks: by treating the dimensional factors multiplied by the dimensionless numbers as constants, the resulting structure reduces to a form akin to an MLP.

The second component, $\mathrm{NeuralOp}_1$, is tasked with approximating the ground-truth operator corresponding to the nondimensionalized PDE in Eq. \eqref{eqn:nondim}. 
By leveraging the similarity transformation invariance property, approximating the operator $\mathcal{G}^{\dagger*}_{\Delta t, \tilde{\mathbf{c}}}(\mathbf{u}^*|_{t_0}) = \mathbf{u_{pred}}^*|_{t_0+\Delta t}$ is equivalent to solving the original problem. From the universal approximation theorem for neural operators, we know this can be achieved by $\mathrm{NeuralOp}_1$.

Finally, the post-processing module $\mathrm{FFW}_{post} \circ \phi_{dim}$ in DimINO can approximate the inverse similarity transformation back to the target physical field. This is ensured by the universal approximation capability of neural networks for general functions.

In summary, the approximation chain can be expressed as:
\begin{equation}
    \begin{split}
        \mathrm{DimINO}(\mathbf{u}) &= \mathrm{FFW}_{post} \circ \phi_{dim} \circ \mathrm{NeuralOp} \\ 
        &\circ \ \mathrm{DimGate}(\mathbf{c}(\mathbf{u}), \ \cdot\ )  \circ \mathrm{FFW}_{pre} \\ 
        &\circ \ \mathrm{LayerNorm} (\mathbf{u}) \\
        \sim \Phi_{post} \ \circ \ & \mathrm{NeuralOp_1} \circ [F_{\alpha}  \circ \mathrm{DimGate}(\mathbf{c}(\mathbf{u}), \ \cdot\ ) \\
        &\circ \ \mathrm{FFW}_{pre} \circ \mathrm{LayerNorm}] (\mathbf{u}) \\
        &= \Phi_{\text{post}, \theta_2} \circ \Phi_{\text{NeuralOp},\theta_1} \circ \Phi_{\text{pre}, \theta_0}.
    \end{split}
\end{equation}

We now present a formal proof sketch of the universal approximation theorem for DimINO, by analyzing the error bounds of its composite function approximation.
Let $\Phi^*_{\text{pre}}$, $\mathcal{G}^{\dagger*}_{\Delta t, \tilde{\mathbf{c}}}$, and $\Phi^*_{\text{post}}$ denote the ideal preprocessing function, the ground-truth evolution operator in the normalized space, and the ideal postprocessing function, respectively. 
We denote $|\cdot| := |\cdot|_{L^2}$ for brevity.
Suppose that the learnable components of the model achieve the following approximation errors:
\begin{align*}
\|\Phi_{\text{pre}, \theta_0}(\mathbf{u}) - \Phi^*_{\text{pre}}(\mathbf{u})\| &\leq \epsilon_1 \\
\|\Phi_{\text{NeuralOp},\theta_1}(\mathbf{u}^*, \tilde{\mathbf{c}}) - \mathcal{G}^{\dagger*}_{\Delta t, \tilde{\mathbf{c}}}(\mathbf{u}^*)\| &\leq \epsilon_2 \\
\|\Phi_{\text{post}, \theta_2}(\mathbf{\hat{u}^*_{pred}}) - \Phi^*_{\text{post}}(\mathbf{\hat{u}^*_{pred}})\| &\leq \epsilon_3
\end{align*}

We decompose the total error using the triangle inequality:
$$
\|\mathcal{G}_\theta(\mathbf{u}) - \mathcal{G}_{\Delta t}(\mathbf{u})\| = $$$$
 \|\Phi_{\text{post}, \theta_2} \circ \Phi_{\text{NeuralOp},\theta_1} \circ \Phi_{\text{pre}, \theta_0}(\mathbf{u}) - \Phi^*_{\text{post}} \circ \mathcal{G}^{\dagger*}_{\Delta t, \tilde{\mathbf{c}}} \circ \Phi^*_{\text{pre}}(\mathbf{u})\|
$$

Adding and subtracting intermediate terms:
$$
\|\mathcal{G}_\theta(\mathbf{u}) - \mathcal{G}_{\Delta t}(\mathbf{u})\| \leq $$$$
 \|\Phi_{\text{post}, \theta_2} \circ \Phi_{\text{NeuralOp},\theta_1} \circ \Phi_{\text{pre}, \theta_0}(\mathbf{u}) - \Phi_{\text{post}, \theta_2} \circ \Phi_{\text{NeuralOp},\theta_1} \circ \Phi^*_{\text{pre}}(\mathbf{u})\| $$$$
\quad + \|\Phi_{\text{post}, \theta_2} \circ \Phi_{\text{NeuralOp},\theta_1} \circ \Phi^*_{\text{pre}}(\mathbf{u}) - \Phi_{\text{post}, \theta_2} \circ \mathcal{G}^{\dagger*}_{\Delta t, \tilde{\mathbf{c}}} \circ \Phi^*_{\text{pre}}(\mathbf{u})\| $$$$
\quad + \|\Phi_{\text{post}, \theta_2} \circ \mathcal{G}^{\dagger*}_{\Delta t, \tilde{\mathbf{c}}} \circ \Phi^*_{\text{pre}}(\mathbf{u}) - \Phi^*_{\text{post}} \circ \mathcal{G}^{\dagger*}_{\Delta t, \tilde{\mathbf{c}}} \circ \Phi^*_{\text{pre}}(\mathbf{u})\|
$$

Since neural networks are Lipschitz continuous on compact domains, we have:
$$
\|\Phi_{\text{NeuralOp},\theta_1}(x_1) - \Phi_{\text{NeuralOp},\theta_1}(x_2)\| \leq L_{\text{neural}} \|x_1 - x_2\| $$$$
\|\Phi_{\text{post}, \theta_2}(y_1) - \Phi_{\text{post}, \theta_2}(y_2)\|\leq L_{\text{post}} \|y_1 - y_2\|
$$

Therefore, each term can be bounded as:
\begin{align*}
\text{Term 1} &\leq L_{\text{post}} L_{\text{neural}} \epsilon_1 \\
\text{Term 2} &\leq L_{\text{post}} \epsilon_2 \\
\text{Term 3} &\leq \epsilon_3
\end{align*}

Combining these bounds:
\begin{equation*}
\|\mathcal{G}_\theta(\mathbf{u}) - \mathcal{G}_{\Delta t}(\mathbf{u})\| \leq L_{\text{post}} L_{\text{neural}} \epsilon_1 + L_{\text{post}} \epsilon_2 + \epsilon_3
\end{equation*}

Given any $\epsilon > 0$, we can choose the network parameters such that:
\begin{align*}
\epsilon_1 &< \frac{\epsilon}{3L_{\text{post}} L_{\text{neural}}}, \quad
\epsilon_2 < \frac{\epsilon}{3L_{\text{post}}}, \quad
\epsilon_3 < \frac{\epsilon}{3}
\end{align*}

This ensures:
\begin{equation*}
d(\mathcal{G}_\theta, \mathcal{G}_{\Delta t}) = \sup_{\mathbf{u}_0\in\mathcal{K}}\|\mathcal{G}_\theta(\mathbf{u}_0) - \mathcal{G}_{\Delta t}(\mathbf{u}_0)\|_{L^2} < \epsilon,
\end{equation*}
which corresponds to Inequation \eqref{neq:operator}.
Therefore, the universal approximation property of DimINO is rigorously established.

\end{proof}

\begin{blackboxtheorem}[Similar Transformation Invariance of DimINO] \label{thm:STI}
Consider any physical PDE system that can be expressed in the form of Eq. \eqref{eqn:pde}, and whose nondimensionalized version is given by Eq. \eqref{eqn:nondim}:

For any two input vectors $\mathbf{u}, \mathbf{u}' \in \mathbb{R}^n$, if the following conditions are satisfied:
\begin{enumerate}
    \item \textbf{Identical dimensionless numbers:} \\$\mathbf{c}(\mathbf{h_{e_0}}(\mathbf{u})) = \mathbf{c}(\mathbf{h_{e_0'}}(\mathbf{u}'))$,
    \item \textbf{Identical dimensionless form:} \\$\mathbf{u}^* = \frac{\mathbf{u}}{\mathbf{h_{e_0}}(\mathbf{u})} = \frac{\mathbf{u}'}{\mathbf{h_{e_0'}}(\mathbf{u'})} = \mathbf{u}'^*$,
\end{enumerate}
then there exists a DimINO model $\mathcal{G}_\theta$ such that:
\begin{enumerate}
    \item \textbf{Same dimensionless prediction:} \\$\mathbf{\hat{u}^*_{pred}}(\mathbf{u}) = \mathbf{\hat{u}^*_{pred}(\mathbf{u}')}$
    \item \textbf{Different redimensionalized outputs:} \\$\mathcal{G}_\theta(\mathbf{u}) = \mathbf{h_{e_0}}(\mathbf{u}) \cdot \mathbf{\hat{u}^*_{pred}} \not \equiv \mathbf{h_{e_0'}}(\mathbf{u'}) \cdot \mathbf{\hat{u}^*_{pred}} = \mathcal{G}_\theta(\mathbf{u}')$
\end{enumerate}
\end{blackboxtheorem}

\begin{proof}
We prove this theorem by analyzing the three-stage architecture of DimINO and showing that each component preserves the similarity transformation invariance.

The preprocessing module $\Phi_{\text{pre}}$ performs dimensionless transformations. 
For inputs $\mathbf{u}$ and $\mathbf{u}'$ satisfying the above conditions, we have:
\begin{align}
\Phi_{\text{pre}}(\mathbf{u}) &= \left(\frac{\mathbf{u}}{\mathbf{h_{e_0}}(\mathbf{u})}, \mathbf{c}(\mathbf{h_{e_0}}(\mathbf{u}))\right) = (\mathbf{u}^*, \mathbf{c}) \\
\Phi_{\text{pre}}(\mathbf{u}') &= \left(\frac{\mathbf{u}'}{\mathbf{h_{e_0'}}(\mathbf{u'})}, \mathbf{c}(\mathbf{h_{e_0'}}(\mathbf{u'}))\right) = (\mathbf{u}'^*, \mathbf{c})
\end{align}
By condition 2: $\mathbf{u}^* = \mathbf{u}'^*$, and by condition 1: $\mathbf{c}(\mathbf{h_{e_0}}(\mathbf{u})) = \mathbf{c}(\mathbf{h_{e_0'}}(\mathbf{u'}))$. 
Therefore:
\begin{equation}
\Phi_{\text{pre}}(\mathbf{u}) = \Phi_{\text{pre}}(\mathbf{u}')
\end{equation}
Since the input to the neural operator blocks is identical for both $\mathbf{u}$ and $\mathbf{u}'$, and these blocks are deterministic, we have:
\begin{equation}
\Phi_{\text{blocks}}(\Phi_{\text{pre}}(\mathbf{u})) = \Phi_{\text{blocks}}(\Phi_{\text{pre}}(\mathbf{u}'))
\end{equation}
Let $\hat{\mathbf{u}}^* = \Phi_{\text{blocks}}(\Phi_{\text{pre}}(\mathbf{u})) = \Phi_{\text{blocks}}(\Phi_{\text{pre}}(\mathbf{u}'))$ be the output from the middle blocks.
The postprocessing module $\Phi_{\text{post}}$ performs redimensionalization by scaling the predicted $\hat{\mathbf{u}}^*$ using the original characteristic scale:
\begin{align}
\mathcal{G}_\theta(\mathbf{u}) &= \Phi_{\text{post}}(\hat{\mathbf{u}}^*) = \mathbf{h}_{e_0}(\mathbf{u}) \cdot \hat{\mathbf{u}}^*_{\text{pred}}, \\
\mathcal{G}_\theta(\mathbf{u}') &= \Phi_{\text{post}}(\hat{\mathbf{u}}^*) = \mathbf{h}_{e_0'}(\mathbf{u}') \cdot \hat{\mathbf{u}}^*_{\text{pred}}.
\end{align}




In summary, the dimensionless predictions $\mathbf{\hat{u}^*_{pred}}$ are identical for physically similar inputs, but the final outputs will differ when different characteristic scales $\mathbf{h_{e_0}}(\mathbf{u}) \neq \mathbf{h_{e_0'}}(\mathbf{u'})$ are used in redimensionalization. 

This proves that DimINO exhibits invariance with respect to similarity transformations in the dimensionless space.

\end{proof}

%% file: contents/3_exp.tex
\section{Experiments}

\begin{table*}[t]
    \centering
    \caption{\textbf{Prediction results}. We report the number of parameters, showing that DimNorm consistently improves performance across all metrics with negligible parameter overhead.
    }
    \vspace{-5pt}
    \label{tab:MajorExperiments}
    \setlength\tabcolsep{8pt}
    \begin{tabular}{lcc|cc|cc|cc}
    \toprule
    \multirow{2}{*}{Dataset} & \multirow{2}{*}{Model} & \multirow{2}{*}{\# Param} & 
    \multicolumn{2}{c|}{rel-L2 ($10^{-2}$)} & \multicolumn{2}{c|}{rel-H1 ($10^{-2}$)} & \multicolumn{2}{c}{rel-L1 ($10^{-2}$)}\\
    &&&score&gain&score&gain&score&gain \\
    \midrule
    
\multirow{2}{*}{Advection1D}
& T-FNO~\citeyearpar{TFNO}  & 51.4K &  1.206 & -     &  4.381 & -     &  1.304 & - \\
& T-FNO + DimINO            & 51.4K &  0.825 & 31.6\%&  3.745 & 37.3\%&  0.854 & 34.5\%  \\
    \midrule
\multirow{2}{*}{Burgers1D}
& T-FNO                     & 51.4K &  0.543 & -     & 1.990  &  -    &  0.491 & - \\
& T-FNO + DimINO            & 51.4K &  0.475 & 12.5\%& 1.885  &  5.3\%&  0.435 & 11.4\%  \\
    \midrule
\multirow{2}{*}{Diffusion-Reaction2D}
& T-FNO                     & 683 K & 24.792 & -     & 26.031 &  -    & 24.837 & - \\
& T-FNO + DimINO            & 684 K & 5.866  & 76.3\%& 6.462  & 75.2\%& 5.523  & 77.8\%  \\
    \midrule
\multirow{2}{*}{TorusVisForce(2D)}
& T-FNO                     & 683 K & 1.203  & -     & 4.227 & -     &  1.166 & -  \\
& T-FNO + DimINO            & 684 K & 0.915  & 23.9\%& 3.838  & 9.2\% &  0.897 & 23.1\%   \\
    \bottomrule
    \end{tabular}
    \vspace{-10pt}
\end{table*}

All experiments are conducted on a server equipped with eight NVIDIA 3080 Ti GPUs.

\subsection{Datasets}
As discussed above, our evaluation focuses on datasets involving multiple physical quantities. Among existing open-source datasets for operator learning, we select the following representative benchmarks: Advection 1D, Burgers 1D, and Diffusion-Reaction 2D from PDEBench~\cite{takamoto2022pdebench}, as well as TorusVisForce~\cite{FFNO} (2D).

\subsection{Choice of Base Models}
Based on preliminary few-shot experiments across these datasets, we identify T-FNO~\cite{TFNO} as the most appropriate baseline. T-FNO is the official successor to the original FNO~\cite{FNO} and is actively maintained by NVIDIA. It consistently outperforms the original FNO in terms of accuracy and eliminates certain resolution-dependent terms present in the original design. 
While other neural operator families, such as CNO~\cite{raonic2024convolutional}, have gained popularity, they suffer from notable limitations: CNO lacks support for 1D datasets and demonstrates poor performance on the 2D benchmarks (often underperforming even the original FNO), while also incurring significantly higher computational costs. Therefore, we adopt T-FNO as our primary baseline model throughout the experiments.

\subsection{Hyperparameters}

In the following experiments, we adopt the optimal configuration recommended in the T-FNO paper.

Rather than using the standard $L2$ loss during training, the authors of T-FNO observed that employing the $H1$ loss leads to faster convergence, even when the evaluation metric remains $L2$. 
We independently verified that this configuration yields superior performance across our tasks and thus adopt it throughout. Notably, we observe consistent improvements without the need for task-specific hyperparameter tuning across different PDE datasets.

We use the default model depth of four layers, as in the original T-FNO implementation. For all PDEBench datasets (Advection 1D, Burgers 1D, and Diffusion-Reaction 2D), we train for a maximum of $1{,}000$ epochs. For the TorusVisForce dataset, we limit training to $500$ epochs.

The time intervals are set to be $T=10$ for the PDE-Bench experiments reported in Table~\ref{tab:MajorExperiments}, and for TorusVisForce experiments, including the major experiment and the Similar Transformation Invariance validation study in Table~\ref{tab:STI}, we set $T=1$ to better accommodate the baseline model’s convergence behavior.

\subsection{Advection}\label{sec:advection}
As a preliminary experiment, we use the 1D Advection dataset from PDEBench as a toy example to evaluate the performance of DimINO.

The advection equation is given as follows, where $\beta$ denotes the advection speed:
$$
\partial_t u(t, x) + \beta \partial_x u(t, x) = 0, x\in (0, 1), t\in (0, 2],
$$
$$
u(0, x) = u_0(x), x\in(0, 1).
$$

This PDE admits an analytical solution:
$$
u(t, x) = u_0(x-\beta t).
$$
The corresponding dimensionless number for this system is $\beta t_0/x_0$. Since we do not vary the characteristic time or length scales in this experiment, we can simply fix $t_0=x_0=1$.

As reported in Table~\ref{tab:MajorExperiments}, DimINO significantly improves over T-FNO on this dataset. Specifically, we observe a $31.6\%$ reduction in relative $L2$ error on the test set.
Similar improvements are observed in both the relative $H1$ and $L1$ errors, indicating consistent performance gains across multiple evaluation metrics.

\subsection{Burgers}\label{sec:burgers}

We further evaluate DimINO on the 1D Burgers dataset from PDEBench~\cite{takamoto2022pdebench} as a representative case for nonlinear dynamics in one-dimensional settings. The Burgers equation is given by: $$\partial_t u(x,t)+u(x,t)\partial_x(u(x,t)) = \nu \partial_{xx}u(x,t), $$$$x\in(0,1). t\in(0,1],$$
where $\nu$ is the viscosity coefficient.

For this system, the relevant dimensionless number is the Reynolds number, defined as $Re = u_0 x_0 / \nu$, where $x_0$ can be set to unit length.

As shown in Table~\ref{tab:MajorExperiments}, DimINO achieves a $12.5\%$ reduction in relative $L2$ error on the test set compared to the baseline. This result further demonstrates the effectiveness of DimINO’s adaptive capabilities in capturing the dynamics of nonlinear PDEs.

\subsection{Diffusion-Reaction 2D}\label{sec:diff-react}

For 2D datasets, we first select the Diffusion-Reaction 2D system from PDEBench~\cite{takamoto2022pdebench}.

The governing equations are given by:
$$
\partial_t u = D_u \partial_{xx}u + D_u \partial_{yy}u + R_u,
$$$$
\partial_t v = D_v \partial_{xx}v + D_v \partial_{yy}v + R_v,
$$
where $D_u$ and $D_v$ are the diffusion coefficients, and $R_u$, $R_v$ represent nonlinear reaction terms.

Diffusion-reaction equations are widely used in modeling chemical systems, though the exact forms of $R_u$ and $R_v$ vary depending on the specific system under consideration.
In PDEBench, the FitzHugh–Nagumo formulation is adopted:
$$
R_u = u-u^3-k-v
$$$$
R_v = u-v.
$$
The associated dimensionless numbers for this system include $D_u/D_v$, $u_0 x_0 / D_u$, and $v_0 x_0 / D_v$, where $x_0$ can be set to unit length.

As shown in Table~\ref{tab:MajorExperiments}, DimINO achieves a substantial improvement of $76.3\%$ in relative $L2$ error over the T-FNO baseline.
We observe that the baseline struggles to converge to a satisfactory error level on this complex system, while the DimINO-enhanced model significantly reduces the error and achieves stable convergence.

\subsection{Navier-Stokes}\label{sec:ns}
For Navier-Stokes problems, we observe that the PDEBench datasets do not include key physical quantities (such as viscosity and external forcing) as inputs, limiting their suitability for evaluating DimINO’s effectiveness. To address this, we instead adopt the TorusVisForce dataset~\cite{FFNO}, which explicitly includes both viscosity and external force terms as part of the input.

TorusVisForce formulates the Navier-Stokes equations in terms of vorticity rather than velocity, thereby reducing the dimensionality of the system. The governing equations are:
\begin{equation}\label{eq:NSvorticity}
    \frac{\partial \omega}{\partial t} + \mathbf{u} \cdot \nabla \omega
    = \nu \nabla^2 \omega + f,
    \
    \nabla \cdot \mathbf{u} = 0,
    \ \nabla \times \mathbf{u} = w \mathbf{e_z}.
\end{equation}
Here, $\omega$ denotes the vorticity, $\nu$ is the viscosity coefficient, and $f$ represents the external forcing. 
The initial input is the vorticity field $\omega_0(x, y) = \omega(x, y, t=0)$, while the force term $f$ remains fixed throughout each sample. Crucially, both $\nu$ and $f$ are randomly sampled during dataset generation and provided as part of the input, ensuring that models must generalize to a variety of physical regimes.

The key dimensionless numbers associated with the vorticity form of the Navier-Stokes equations include: (1) Reynolds number: $Re = \frac{w_0 L^2}{\nu}$; (2) Strouhal Number $St = {t_0 w_0}$; (3) Froude Number $Fr = w_0\sqrt{\frac{L}{f_0}}$.

In this experiment, we use $T=10$ for the prediction time step. As shown in the fourth row of Table~\ref{tab:MajorExperiments}, DimINO consistently outperforms the baseline T-FNO across all evaluation metrics on the TorusVisForce dataset.

\subsection{Validation for Similar Transformation Invariance}
\label{exp:STI}

We further evaluate DimINO’s invariance to similar transformations on one of the most challenging tasks: TorusVisForce. This dataset provides an ideal testbed, as its governing equation (Eq.~\ref{eq:NSvorticity}) shares the same dimensional structure as the general Navier-Stokes equations discussed earlier.

In this setting, we define the characteristic quantities as follows: (1) Characteristic length $L$, (2) Characteristic vorticity of fluid $w_0$, (3) Characteristic time interval $t_0$, and (4) Characteristic forcing $f_0$. 
Applying dimensional analysis leads to the following nondimensionalization: 
$$
w'=\frac{w}{w_0}, \boldsymbol{r'} = \frac{\boldsymbol{r}}{L}, t' = \frac{t}{t_0}, p' = \frac{p}{\rho U^2}, f' = \frac{f}{f_0}
$$
Substituting into Eq.~\eqref{eq:NSvorticity}, we obtain the dimensionless vorticity equation:
$$
\frac{1}{St} (\frac{d}{dt'} + \boldsymbol{v}\cdot \nabla') w' = \frac{1}{Re} \nabla'^2 w' + \frac{1}{Fr^2} f',
$$
where the dimensionless numbers remain the same as mentioned in the previous experiments.

In TorusVisForce, the characteristic length $L$ is always fixed to the unit domain length, so affine transformations are primarily applied to vorticity, forcing, and time. 
In particular, we consider the prediction time interval $T$ as the characteristic time $t_0$, assuming it remains within the dominant timescale of the fluid’s wave dynamics. Accordingly, we set $T=1$ for this study.

To implement a similar transformation in practice, we apply the following input scaling:
$$
    \omega_0 \rightarrow \omega_0/p,\quad t \rightarrow t\cdot p,\quad  \nu \rightarrow \nu/p, \quad f \rightarrow f/p^2.
$$
Under this transformation, the predicted output is expected to satisfy:
$$
    \omega_\text{new}(x, T_\text{new}=pT)=k \omega(x, T),
$$
which is verified using a traditional numerical solver, up to minor numerical errors.

For DimINO, such invariance can be captured implicitly. From the untransformed task perspective, achieving this behavior would require the model to roll out predictions $p$ times over horizon $T$.
However, DimINO can directly adapt to the scaled Strouhal number, preserving predictive accuracy under transformation.

As shown in Table~\ref{tab:STI}, the relative $L2$ error of DimINO remains nearly unchanged under similar transformations, indicating strong invariance. 
In contrast, the pretrained T-FNO baseline accumulates significant error over $p$ rollouts, and the transformed inputs become severely out-of-distribution for the original model. These findings demonstrate that DimINO effectively captures the underlying invariance in transformed physical regimes, which explains its markedly better performance compared to the baseline.

\begin{table}[t]
    \centering
    \caption{\textbf{Validation on the STI property.} We conduct experiments on TorusVisForce ($T=1$). 
    }
    \vspace{-5pt}
    \label{tab:STI}
    \setlength\tabcolsep{5pt}
    \begin{tabular}{lrrrr}
    \toprule
    \multirow{2}{*}{Model}  & \multicolumn{4}{c}{rel-L2 ($10^{-2}$)} \\
                            & p=1 & p=2 & p=4 & p=8 \\
    \midrule
    T-FNO & 1.203  & 23.937 & 60.145 & 79.986 \\
    T-FNO + DimINO        & 0.915  & 0.915  & 0.915  & 0.915  \\
    Gain                     & 23.9\% & 96.2\% & 98.5\% & 98.9\% \\

    \bottomrule
    \end{tabular}
\end{table}

%% file: contents/5_concl.tex
\section{Conclusions and Limitations}

In this paper, we introduce DimINO, a design methodology for modeling high-dimensional PDEs by embedding physical dimensional relationships through DimNorm and redimensionalization.
This approach benefits both interpretability and learning, especially when models take physical quantities as inputs. 
Beyond a specific technique, DimINO offers a new design axis for AI4Science models by using physical dimensionality as a guiding principle. It achieves up to $76.3\%$ performance improvement on PDE benchmarks.
To address the open challenge of quantifying the benefits of encoding dimensional properties, we propose evaluating Similar Transformation Invariance (STI) as a diagnostic tool for generalization. Results show that incorporating dimensional priors significantly improves STI robustness.

A potential limitation of DimINO lies in tasks where input data lacks meaningful dimensional structure. In such cases, the benefits may be diminished, as DimNorm may behave similarly to standard normalization layers like LayerNorm.